\documentclass{article}

\usepackage{microtype}
\usepackage{graphicx}
\usepackage{booktabs} % for professional tables
\usepackage{float} 
\usepackage{wrapfig}
\usepackage{caption}
\usepackage{subcaption}
\usepackage{wrapfig}
\usepackage{bbm}
\usepackage{hyperref}

\usepackage[accepted]{icml2024}

\usepackage{amsmath}
\usepackage{amssymb}
\usepackage{mathtools}
\usepackage{amsthm}

\usepackage[capitalize,noabbrev]{cleveref}

\theoremstyle{plain}

\theoremstyle{definition}

\theoremstyle{remark}

\usepackage[textsize=tiny]{todonotes}

\icmltitlerunning{ConvNet vs Transformer, Supervised vs CLIP: Beyond ImageNet Accuracy}
\newcommand{\cnnsup}{ConvNeXt-sup}
\newcommand{\cnnclip}{ConvNeXt-clip}
\newcommand{\vitsup}{ViT-sup}
\newcommand{\vitclip}{ViT-clip}

\newlength\savewidth\newcommand\shline{\noalign{\global\savewidth\arrayrulewidth\global\arrayrulewidth1pt}\hline\noalign{\global\arrayrulewidth\savewidth}}

\begin{document}

\twocolumn[
\icmltitle{ConvNet vs Transformer, Supervised vs CLIP: Beyond ImageNet Accuracy}



\icmlsetsymbol{equal}{*}

\begin{icmlauthorlist}
\icmlauthor{Kirill Vishniakov}{mbz}
\icmlauthor{Zhiqiang Shen}{mbz}
\icmlauthor{Zhuang Liu}{meta}

\end{icmlauthorlist}

\icmlaffiliation{mbz}{MBZUAI}
\icmlaffiliation{meta}{Meta AI Research}

\icmlcorrespondingauthor{Kirill Vishniakov}{ki.vishniakov@gmail.com}

\icmlkeywords{Machine Learning, ICML}

\vskip 0.3in
]



\printAffiliationsAndNotice{} % otherwise use the standard text.

\begin{abstract}

 Modern computer vision offers a great variety of models to practitioners, and selecting a model from multiple options for specific applications can be challenging. Conventionally, competing model architectures and training protocols are compared by their classification accuracy on ImageNet. However, this single metric does not fully capture performance nuances critical for specialized tasks. In this work, we conduct an in-depth comparative analysis of model behaviors beyond ImageNet accuracy, for both ConvNet and Vision Transformer architectures, each across supervised and CLIP training paradigms. Although our selected models have similar ImageNet accuracies and compute requirements, we find that they differ in many other aspects: types of mistakes, output calibration, transferability, and feature invariance, among others. This diversity in model characteristics, not captured by traditional metrics, highlights the need for more nuanced analysis when choosing among different models. Code is available at \href{https://github.com/kirill-vish/Beyond-INet}{github.com/kirill-vish/Beyond-INet}.
\end{abstract}  

\section{Introduction}

The computer vision model landscape has become increasingly complex. From early ConvNets~\citep{lecun1998gradient} to advances in Vision Transformers~\citep{dosovitskiy2020image}, the variety of models available has expanded significantly. Similarly, training paradigms have evolved from supervised training on ImageNet~\citep{inet_dataset} to self-supervised learning~\citep{chen2020simple, he2020momentum} and image-text pair training like CLIP~\citep{radford2021learning}. While signaling progress, this explosion of choices poses a significant challenge for practitioners: how to select a model that suits their purposes?

\begin{figure}
\centering
\includegraphics[width=1\linewidth]{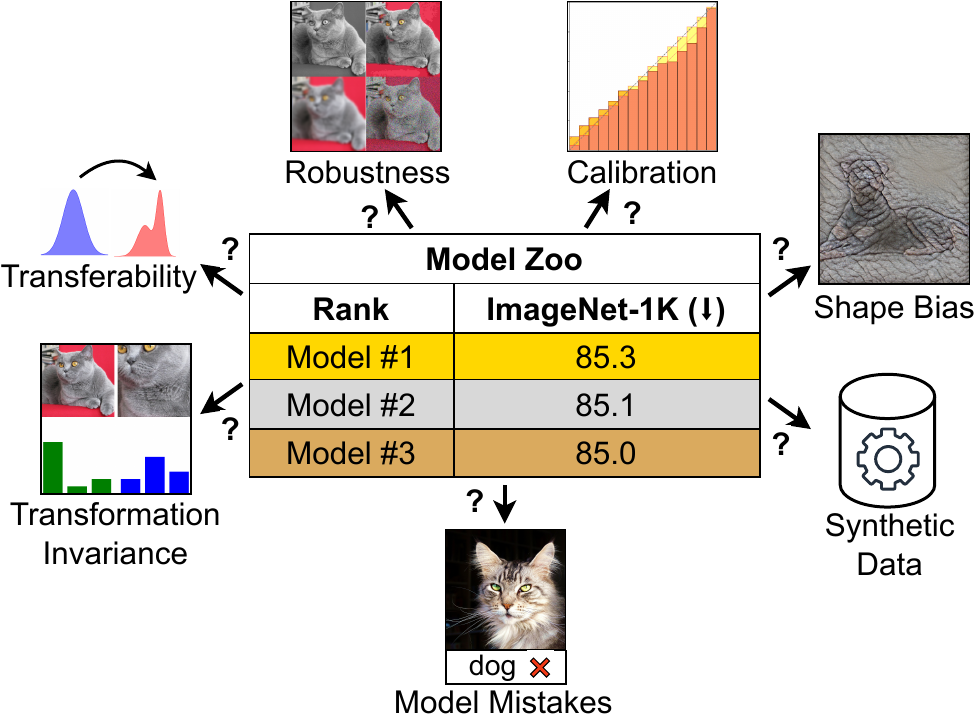}
\vspace*{-0.25in}
\caption{Models are often compared only by their ImageNet accuracy, without looking at many other important behaviors. In our work, we analyze models with similar ImageNet accuracies and find that they have vastly different properties.}
\label{fig:fig1}
\vspace*{-0.3in}
\end{figure}

Conventionally, ImageNet accuracy has served as the primary metric for evaluating model performance. It has driven remarkable progress since it ignited the deep learning revolution \citep{NIPS2012_4824}. However, this metric is becoming increasingly insufficient. While ImageNet is useful to measure a model's general capability, it does not capture the nuanced differences arising from varying architectures, training paradigms, and data -- models with different properties may appear similar if judged solely based on ImageNet accuracy (Fig.~\ref{fig:fig1}). This limitation becomes more pronounced as models start to overfit the idiosyncrasies of ImageNet with saturated accuracies~\citep{beyer2020we}.

A particularly noteworthy example is CLIP. Despite having a similar ImageNet accuracy as a ResNet~\citep{he2016deep}, CLIP's vision encoder exhibits much better robustness and transferability. This has sparked research that explores and builds upon the unique strengths of CLIP~\citep{ramesh2022hierarchical, luo2022clip4clip, wortsman2022robust, vinker2023clipascene}, which were not evident from the ImageNet metric alone. This demonstrates that analyzing alternative properties could help discover useful models.
\begin{table*}[!htbp]
    \vspace*{-0.05in}
    \scriptsize
    \addtolength{\tabcolsep}{0pt}
    \def\arraystretch{1.2}
    \centering
    \resizebox{0.99\textwidth}{!}{
    \begin{tabular}{lccccccc}
        \bf Model & \bf Architecture & \bf Pretraining & \bf Finetuning & \bf Paradigm & \bf FLOPs & \bf \#Param & \bf INet-1K val\% \\
        \shline
        \vitsup & ViT-B/16 & ImageNet-21K & ImageNet-1K & supervised &  17.5G & 87M & 85.5 \\
        \cnnsup & ConvNeXt-B & ImageNet-21K & ImageNet-1K & supervised &  15.4G & 89M & 85.5 \\ 
        \vitclip & ViT-B/16 &  LAION-400M & --- & CLIP & 17.5G & 87M & 67.0 \\
        \cnnclip & ConvNeXt-B & LAION-400M & --- & CLIP & 15.4G & 89M & 66.3 \\
    \end{tabular}
    }
    \vspace{-0.1in}
    \caption{\textbf{Model summary in our analysis.} We select ConvNeXt and ViT with similar ImageNet accuracies within each training paradigm.}
    \label{tab:model_details}
    \vspace{-0.15in}
\end{table*}

In addition to fundamental research, the growing integration of vision models into production systems also calls for a deep understanding of their behaviors. Conventional metrics do not fully capture models' ability to handle real-world vision challenges like varying camera poses, lighting conditions, or occlusions. For instance, models trained on datasets such as ImageNet often struggle~\citep{yamada2022does} to transfer their performance to real-world applications where conditions and scenarios are more diverse.

To bridge this gap, we conduct an in-depth exploration focusing on model behaviors beyond ImageNet accuracy. We analyze four leading models in the computer vision: ConvNeXt~\citep{liu2022convnet}, as a representative ConvNet, and Vision Transformer (ViT)~\citep{dosovitskiy2020image}, each under supervised and CLIP training. The selected models are similar in parameter counts and show nearly identical accuracy on ImageNet-1K within each training paradigm, ensuring a fair comparison. Our study delves into a wide array of model characteristics, such as types of prediction errors, generalization capabilities, invariances of the learned representations, calibration, and many others. Importantly, our focus is on properties exhibited by the model without additional training or finetuning, providing insights for practitioners interested in using pretrained models directly.

In our analysis, we discover substantial variations in model behaviors among different architectures and training paradigms. For example, CLIP models make fewer classification errors relative to their ImageNet performance. However, supervised models are better calibrated and superior on ImageNet robustness benchmarks. ConvNeXt has an advantage on synthetic data but is more texture-biased than ViT. We also find that supervised ConvNeXt excels on many benchmarks and achieves transferability comparable to that of CLIP models. Based on these findings, it becomes evident that various models demonstrate their strengths in unique ways that are not captured by a single metric. Our research emphasizes the need for more detailed evaluation metrics for accurate, context-specific model selection and the creation of new benchmarks unrelated to ImageNet.

\vspace{-0.05in}
\section{Models}

For analyzing ConvNets and Transformers, many previous works~\citep{naseer2021intriguing, minderer2021revisiting, zhou2021convnets, NEURIPS2021_e19347e1} compare ResNet and ViT. This comparison is often disadvantageous for ConvNet since ViTs are typically trained with more advanced recipes, achieving higher ImageNet accuracy. ViT also has architecture design elements, e.g., LayerNorm~\citep{ba2016layer}, that were not incorporated in ResNet when it was invented years ago. For a more balanced evaluation, we compare ViT with ConvNeXt~\citep{liu2022convnet}, a modern of ConvNet that matches Transformers' performance.

As for the training paradigms, we compare supervised and CLIP. Supervised models continue to show state-of-the-art performance in computer vision~\citep{dehghani2023scaling}. CLIP models, on the other hand, excel in generalization and transferability, and offer intriguing representational properties that connect vision and language. Self-supervised models~\citep{he2022masked,woo2023convnext} are not included in the results as they showed behaviors similar to supervised models in our preliminary tests. This could be due to their final ImageNet-1K supervised finetuning, which is necessary for studying many properties.

The selected models have similar ImageNet-1K validation accuracies within their respective training paradigms, ensuring a fair comparison. For CLIP models, these indicate their zero-shot accuracies. The models also have similar sizes and computational requirements, and are publicly available. Since we are using pretrained models, we cannot control for the number and quality of data samples seen during training. 

For supervised models, we use a pretrained DeiT3-Base/16~\citep{touvron2022deit} for ViT, which shares the same architecture as ViT-Base/16 with an improved training recipe, and ConvNeXt-Base~\citep{liu2022convnet}. For CLIP models, we use vision encoders of ViT-Base/16 and ConvNeXt-Base from OpenCLIP~\citep{ilharco_gabriel_2021_5143773}. Note that these models have a slightly different performance from the original OpenAI models~\citep{radford2021learning}. A detailed model comparison is given in Table~\ref{tab:model_details}. 

We recognize there are other ViT CLIP models pretrained on larger datasets such as LAION-2B, DataComp~\cite{gadre2023datacomp}, DFN~\cite{fang2023data}, which show a better performance. However, OpenCLIP offers only a few pretrained ConvNeXt models and for most of them there is no matching ViT counterpart in terms of ImageNet accuracy, pretraining dataset and  parameter count. Therefore, we chose CLIP models pretrained on LAION-400M, offering the most fair comparison between ConvNet and ViT.   

\textbf{Additional models of different sizes.} In the Appendix~\ref{appendix:models}, we present the performance results for models of different sizes to evaluate the impact of model size on performance. These models are assessed across several benchmarks, including ImageNet-X, PUG-ImageNet, calibration, invariance, and shape/texture bias.

\vspace{-0.05in}
\begin{figure*}[ht]
    \centering
    \includegraphics[width=\textwidth]{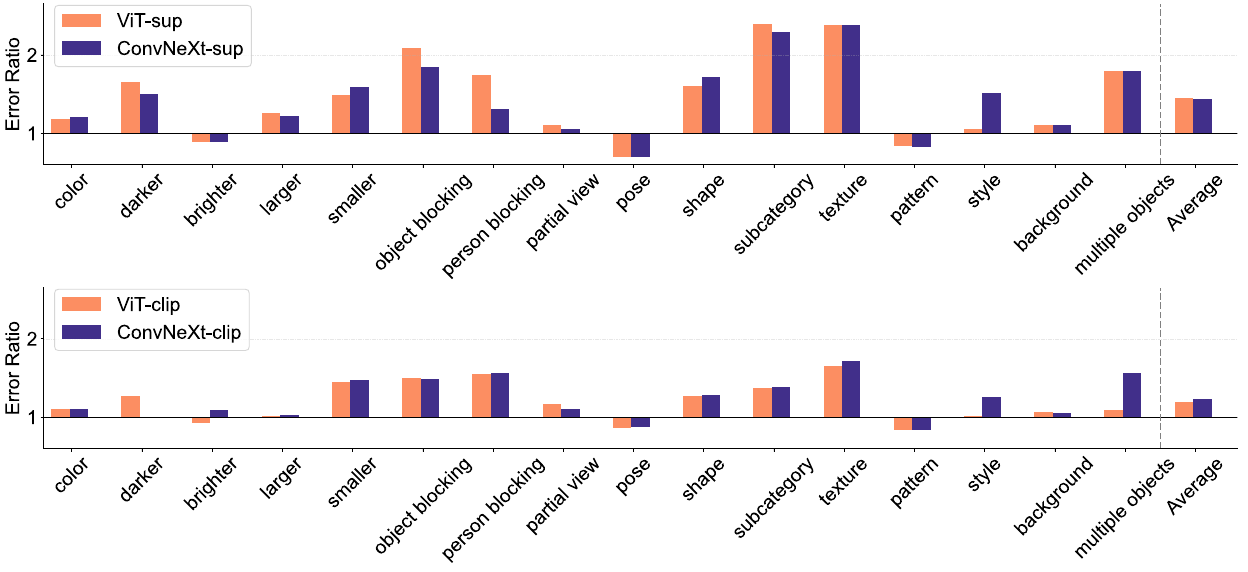}
  \vspace{-0.3in}
  \caption{\textbf{Model mistakes on ImageNet-X.} Lower is better. ConvNeXt and ViT perform similarly within each training category. CLIP models achieve lower error ratios compared to supervised.}
  \label{fig:imagenetx}
  \vspace{-0.125in}
\end{figure*}

\section{Property Analysis}

Our analysis is designed to investigate model behaviors that can be evaluated without the need for further training or finetuning. This approach is particularly relevant for practitioners with limited compute resources, who often rely on pretrained models. While we recognize the value of downstream tasks like object detection, our focus is on properties that offer insights with minimal computational demands and reflect behaviors important for real-world applications. 

\vspace{-0.05in}
\subsection{Model Mistakes}

 In image classification, a model mistake is an incorrect label assignment, such as misclassifying a cat as a dog. Simply identifying mistaken object classes might not offer actionable insights for model improvement. The key aspect, therefore, is finding the specific reasons for these mistakes. For instance, some models may be particularly sensitive to certain aspects of the data distribution, like texture variations. In this case, a model might consistently make mistakes when the texture of the object differs from what it has been trained on. Identifying mistake types allows for targeted data collection and retraining, improving over a black-box approach.

The ImageNet-X dataset~\citep{idrissi2022imagenet} offers detailed human annotations for 16 factors of variation, such as pose, style, and others. This allows a focused analysis of models' mistake types. The annotations enable measuring model error ratios for each factor independently: $\text{error ratio(factor)} = \frac{1 - \text{accuracy(factor)}}{1 - \text{accuracy(overall)}},$
where $\text{accuracy(overall)}$ is the overall ImageNet-1K validation accuracy, and $\text{accuracy(factor)}$ is the accuracy on all the images where the factor was highlighted. This metric measures the model performance on a given factor relative to its overall performance. Lower error ratios indicate better performance, implying higher accuracy for the specific factor. Our results on ImageNet-X are presented in Fig.~\ref{fig:imagenetx}.

\textbf{CLIP models make fewer mistakes relative to their ImageNet accuracy than supervised.} The diagram in Fig.~\ref{fig:imagenetx} shows that CLIP models have a smaller error ratio, indicating a significant advantage over supervised models. However, it is important to note that the error ratio is relative to overall ImageNet accuracy, where a significant 18\% gap exists between supervised and CLIP zero-shot models. In particular, CLIP models are much more robust towards shape, subcategory, texture, object blocking, and darker factors. The key reason for the success of CLIP models is likely the more diverse data used for training.

\textbf{All models suffer mostly from complex factors like occlusion.} For CLIP models, there are three factors with dissimilar performance between ConvNeXt and ViT: multiple objects, style, and darker. For the first two, the ConvNeXt has a higher error ratio, while for the latter, it has an advantage over ViT. For supervised models, the performance only diverges for style and person blocking. Except for these factors, models largely have similar error ratios. The six factors for which all the models have a high error ratio are smaller, object blocking, person blocking, shape, subcategory, and texture. High error ratio factors usually involve complex visual scenarios, which helps to explain why models often make mistakes in these situations. For example, in cases of occlusion, the model often misclassifies due to focusing on the visible, obscuring object.
\begin{figure*}
    \centering
    \begin{minipage}[b]{0.66\textwidth}  
        \centering
        \begin{minipage}[b]{0.49\textwidth}
            \includegraphics[width=1\textwidth]{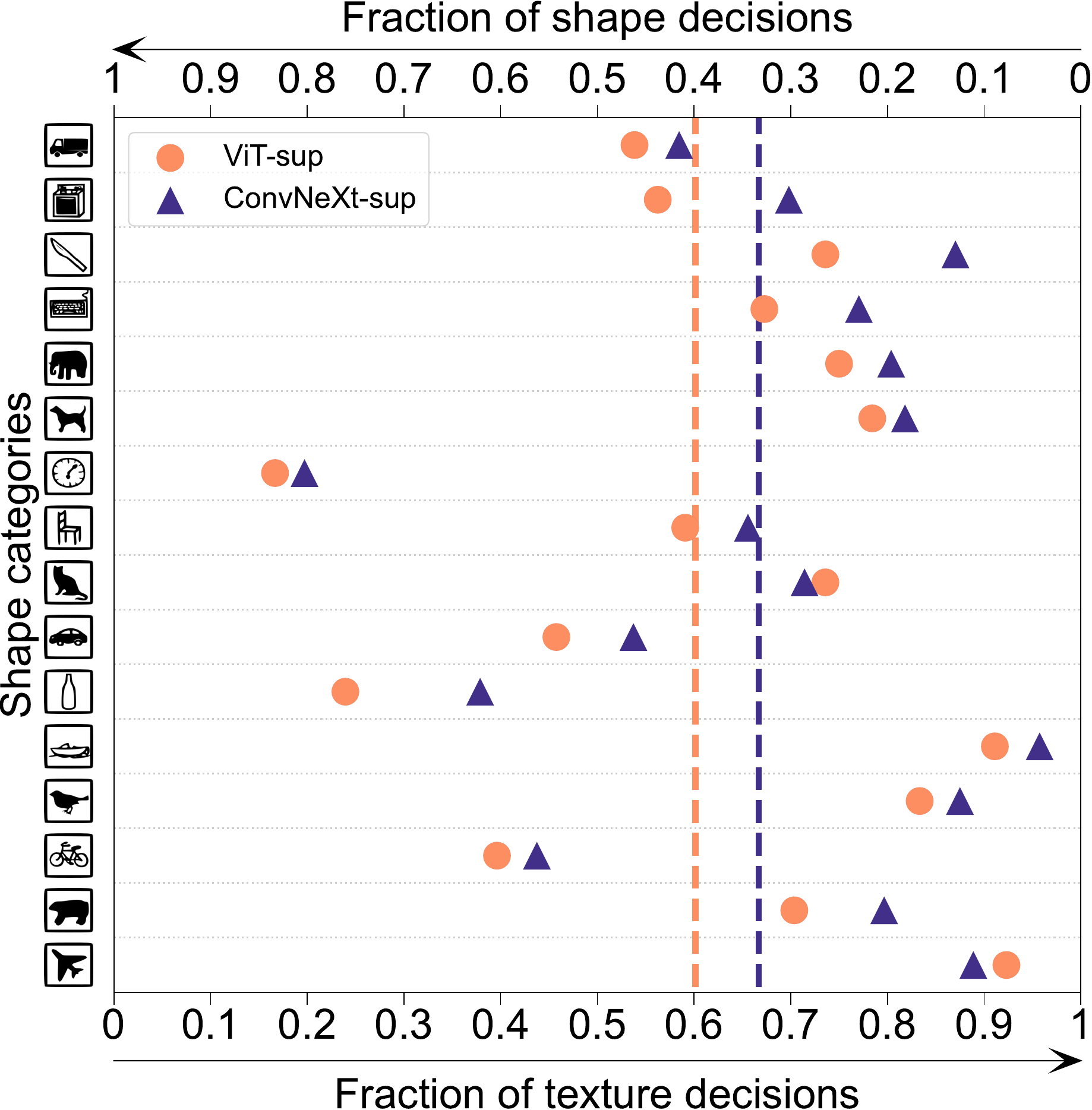}
        \end{minipage}
        \begin{minipage}[b]{0.49\textwidth}
            \includegraphics[width=1\textwidth]{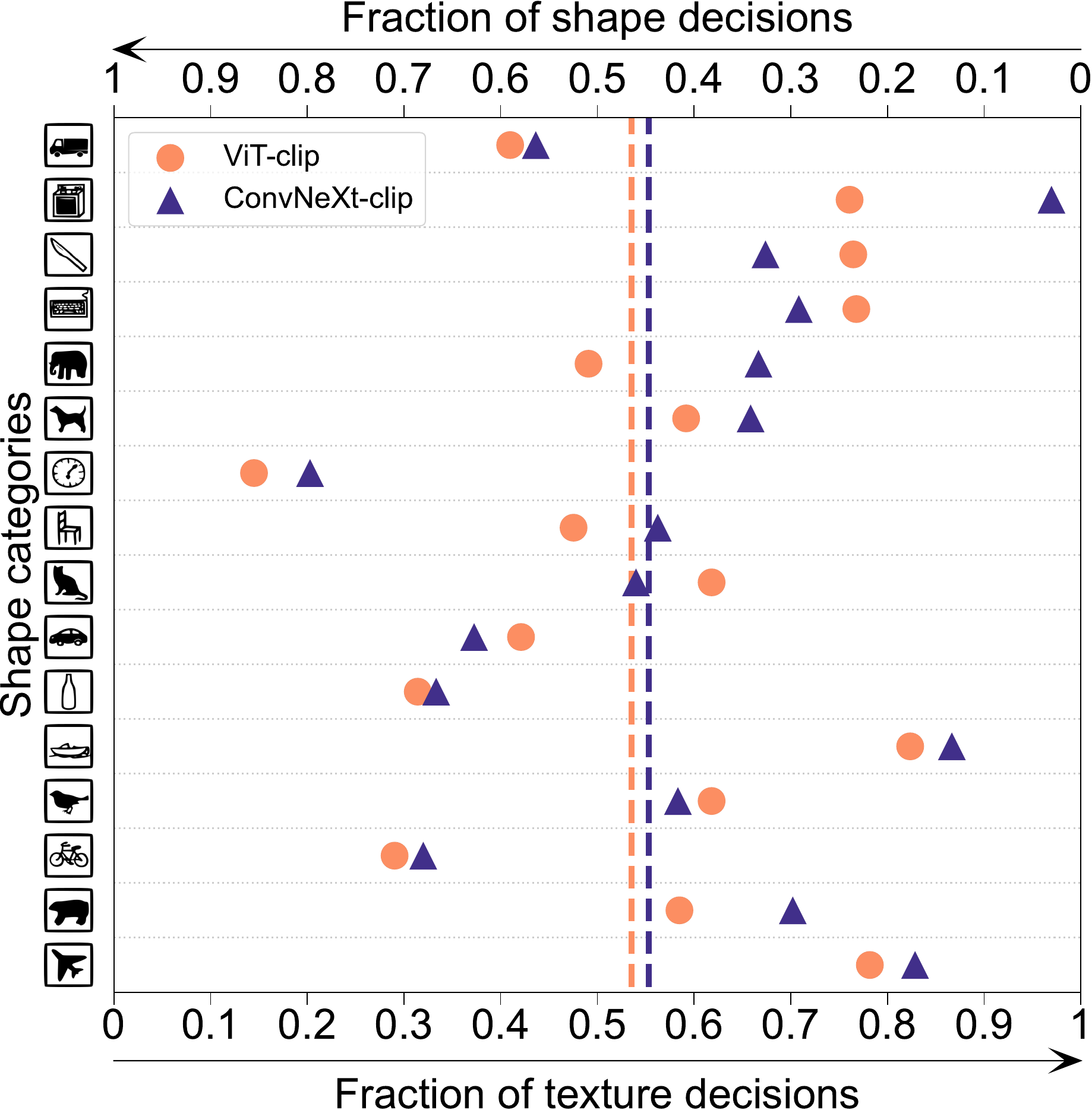}
        \end{minipage}
        \caption{\textbf{Fraction of shape vs texture decisions on cue-conflict dataset.} ViT models show a higher shape bias. CLIP models are less texture-biased than their supervised counterparts. All models still have a significant fraction of texture decisions.}
        \label{fig:shape_texture_bias}
    \end{minipage}\hfill 
    \begin{minipage}[b]{0.29\textwidth}
        \raisebox{0.14\textwidth}{
            \includegraphics[width=0.98\textwidth]{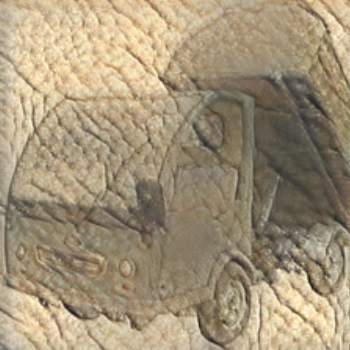}
        }
        \raisebox{0.15\textwidth}{
            \parbox{\linewidth}{
                \caption{A cue-conflict image~\citep{geirhos2018imagenet}.}
                \label{fig:cue_conflict}
            }
        }
    \end{minipage}
    \vspace{-0.152in}
\end{figure*}

\textbf{Texture is the most challenging factor for all models.}
Interestingly, all models in our analysis have the largest error ratio on the texture factor. It refers to images where the texture of the object differs from its standard appearance. This suggests that models of the current generation largely suffer because of texture bias. 

\subsection{Shape / Texture Bias}
\label{section:shape_texture}

In contrast to humans, who generally use high-level visual cues for recognition, neural networks often rely on more brittle shortcut features~\citep{geirhos2020shortcut}. The study of shape-texture bias~\citep{geirhos2018imagenet} serves to highlight this phenomenon by examining model behavior on cue-conflict images, which contain a shape from one class superimposed with the texture from another (Fig. \ref{fig:cue_conflict}). Two key metrics are introduced to quantify this bias: the shape and the texture fractions. The shape fraction calculates the proportion of decisions leaning towards the class represented by the shape, while the texture fraction measures those for the texture class. These metrics reveal whether the classifier favors shape or texture when they conflict.

The study in~\cite{geirhos2018imagenet} showed that ConvNets have a strong bias towards texture, as opposed to shape, which differs from humans. Subsequent work~\cite{naseer2021intriguing} concluded that ViT is less biased towards the texture than ConvNet by comparing the first generation of DeiT-S~\citep{touvron2021training} and ResNet-50. Remarkably, scaling large Transformer models has led to shape biases comparable to human level~\citep{dehghani2023scaling}. 

We evaluate shape-texture bias in our models using cue-conflict images and display the findings in Fig.~\ref{fig:shape_texture_bias}. Dashed lines represent average shape bias aggregated over all the categories. Individual markers on horizontal lines depict shape bias for the particular class, which is identified by a corresponding logo on the y-axis. The shape fraction is represented on the top x-axis of the diagrams, while the bottom x-axis indicates the texture fraction.

\textbf{CLIP models have smaller texture bias than supervised.} In Fig.~\ref{fig:shape_texture_bias}, we can observe that ViTs exhibit stronger shape bias than ConvNeXts for both supervised and CLIP models. This is possibly because ConvNeXt is more inclined to learn local features related to textures due to the local nature of convolution operation. However, the gap between ViT and ConvNeXt is much smaller for CLIP-based models. Notably, the shape bias in CLIP models improved by 7\% and 12\% for both architectures, prompting questions about the benefits of further scaling the training data. ConvNets typically exhibit lower shape bias compared to ViT, however, the gap for CLIP models is marginal. In~\cite{dehghani2023scaling}, it has been shown that a 22B parameter ViT model can achieve 87\% shape bias. In our analysis, the ViT CLIP model achieved a maximum shape bias of 46.4\%, suggesting that the model size might also play an important role.

\subsection{Model Calibration}
\begin{figure*}[ht]
    \vspace*{-0.06in}
    \centering
    \begin{subfigure}{0.999\textwidth}
        \centering
        \includegraphics[width=\textwidth]{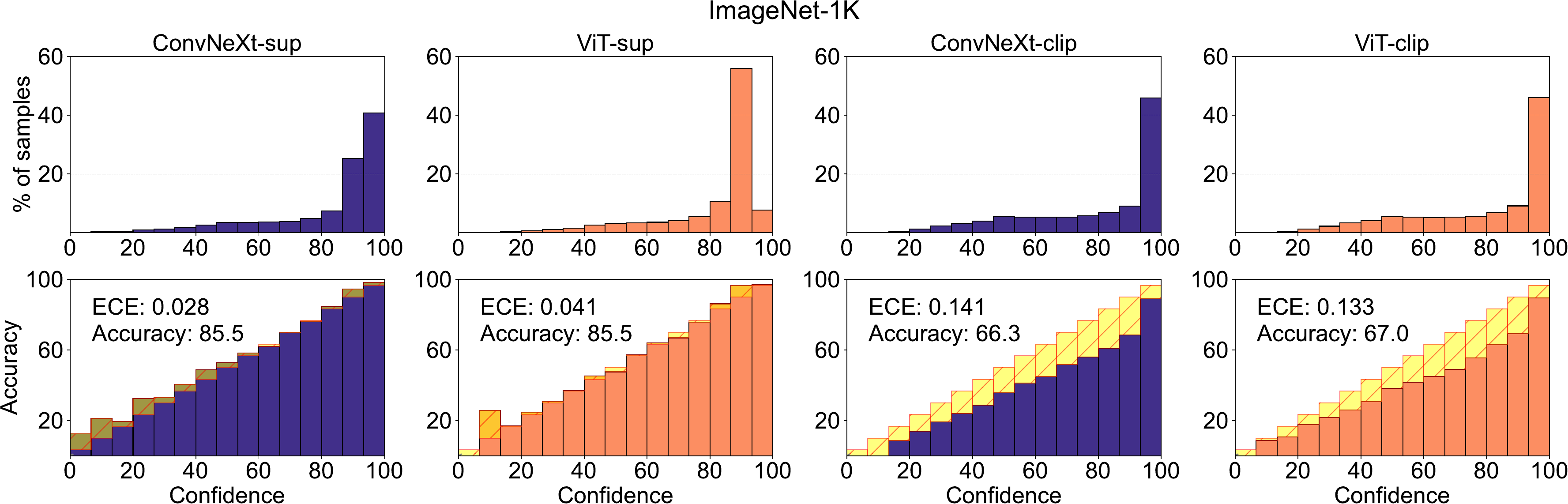}
    \end{subfigure}
    \vspace*{0.05in}
    \begin{subfigure}{0.999\textwidth}
        \centering
        \includegraphics[width=\textwidth]{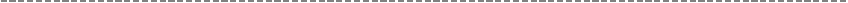}
    \end{subfigure}
    \begin{subfigure}{0.999\textwidth}
        \centering
        \includegraphics[width=\textwidth]{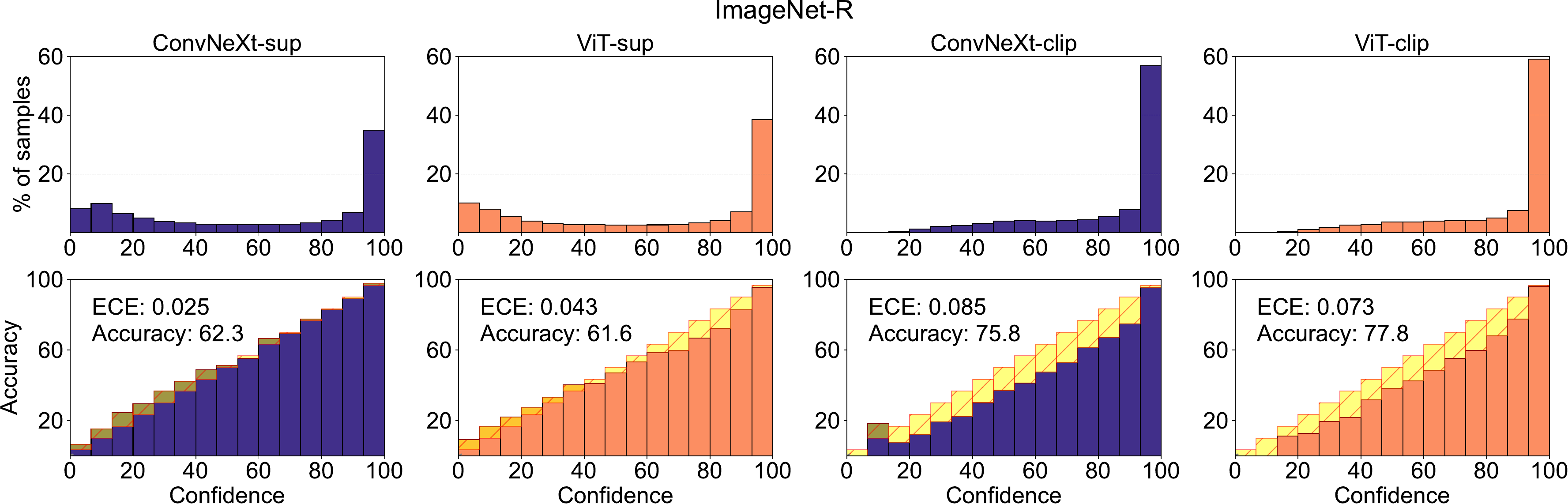}
    \end{subfigure}
    \vspace{-0.3in}
    \caption{\textbf{Calibration results}: confidence histograms (1 and 3 row), reliability diagrams (2 and 4 row), and ECE on ImageNet-1K (top) and ImageNet-R (bottom). Supervised models have lower ECE in both cases. CLIP models have bars under diagonal and many high confidence predictions, indicating overconfidence. ConvNeXt is better (ImageNet-1K) or competitive (ImageNet-R) to ViT.}
    \label{fig:calibration}
    \vspace{-0.1225in}
\end{figure*}
Besides vulnerability to shortcut features, poor model performance can often be attributed to miscalibration, where a model's confidence in its predictions does not align with actual accuracy. Model calibration is a metric that quantifies the reliability of a model's predicted confidence levels~\citep{guo2017calibration}. 
A model's confidence for a prediction is defined as the max probability among all classes in its output distribution. 
We are interested in determining whether the model is overly confident or too uncertain in its predictions. For instance, if the network deems a set of predictions to be 80\% confident, does the actual accuracy hover around 80\%?

The calibration rate can be quantified by Expected Calibration Error (ECE). To calculate ECE, predictions first need to be separated into the $M$ bins $B_1, \dotso, B_M$ based on their confidence. For instance, one bin can include all the predictions with confidence between 50\% and 60\% and so on. Each bin's confidence and accuracy are calculated as the average confidence and accuracy of predictions in $B_i$, represented as $\text{conf}(B_i)$ and $\text{acc}(B_i)$. Then, ECE can be defined as:
$
\text{ECE} = \sum_{i}^{M} \frac{| B_i |}{n} \left| \text{acc}(B_i) - \text{conf}(B_i) \right|, 
\label{eq:ece}
$
where $|B_i|$ is the size of the $i$-th bin.

Model calibration is also often assessed through visualizations, including reliability diagrams and confidence histograms. Reliability diagrams plot the predicted confidence against accuracy; a well-calibrated model would show a graph where points closely align with the diagonal. Confidence histograms display how often different confidence levels occur in the model's predictions.

For a balanced evaluation, we present calibration metrics on two different datasets: ImageNet-1K for in-distribution data and ImageNet-R~\citep{hendrycks2021many} for out-of-distribution data. We select ImageNet-R as the out-of-distribution because CLIP models show higher accuracy on it than supervised. In all experiments, we use $M=15$ bins. We plot confidence histograms (1 and 3 rows), reliability diagrams (2 and 4 rows), and ECE in Fig.~\ref{fig:calibration}.

\textbf{CLIP models are overconfident and supervised models are slightly underconfident.} In Fig.~\ref{fig:calibration}, we observe that CLIP models have bars consistently below the diagonal in reliability diagrams and a notably high last bar in the confidence histogram, signaling overconfidence in both in-distribution and out-of-distribution data. Although \cite{minderer2021revisiting} attributes calibration performance mainly to architecture, our results suggest otherwise: higher ECE scores in CLIP models, despite superior accuracy on ImageNet-R, indicate that training data and objectives could be more influential factors. We also highlight that our results are different from \citep{minderer2021revisiting} for CLIP models presumably because they use checkpoints from OpenAI~\citep{radford2021learning} and we use from OpenCLIP~\citep{ilharco_gabriel_2021_5143773}.
In the lower part of Fig.~\ref{fig:calibration} related to ImageNet-R, we note that supervised models exhibit a higher density in the lower confidence intervals of the confidence histograms (3rd row). Moreover, these models show elevated accuracy levels in the initial bins of the reliability diagrams (4th row). These findings suggest that supervised models tend to be slightly underconfident on ImageNet-R.

\begin{figure*}[h]
    \centering
    \vspace*{-0.1in}
    \begin{minipage}[b]{0.499\textwidth}
        \includegraphics[width=\textwidth]{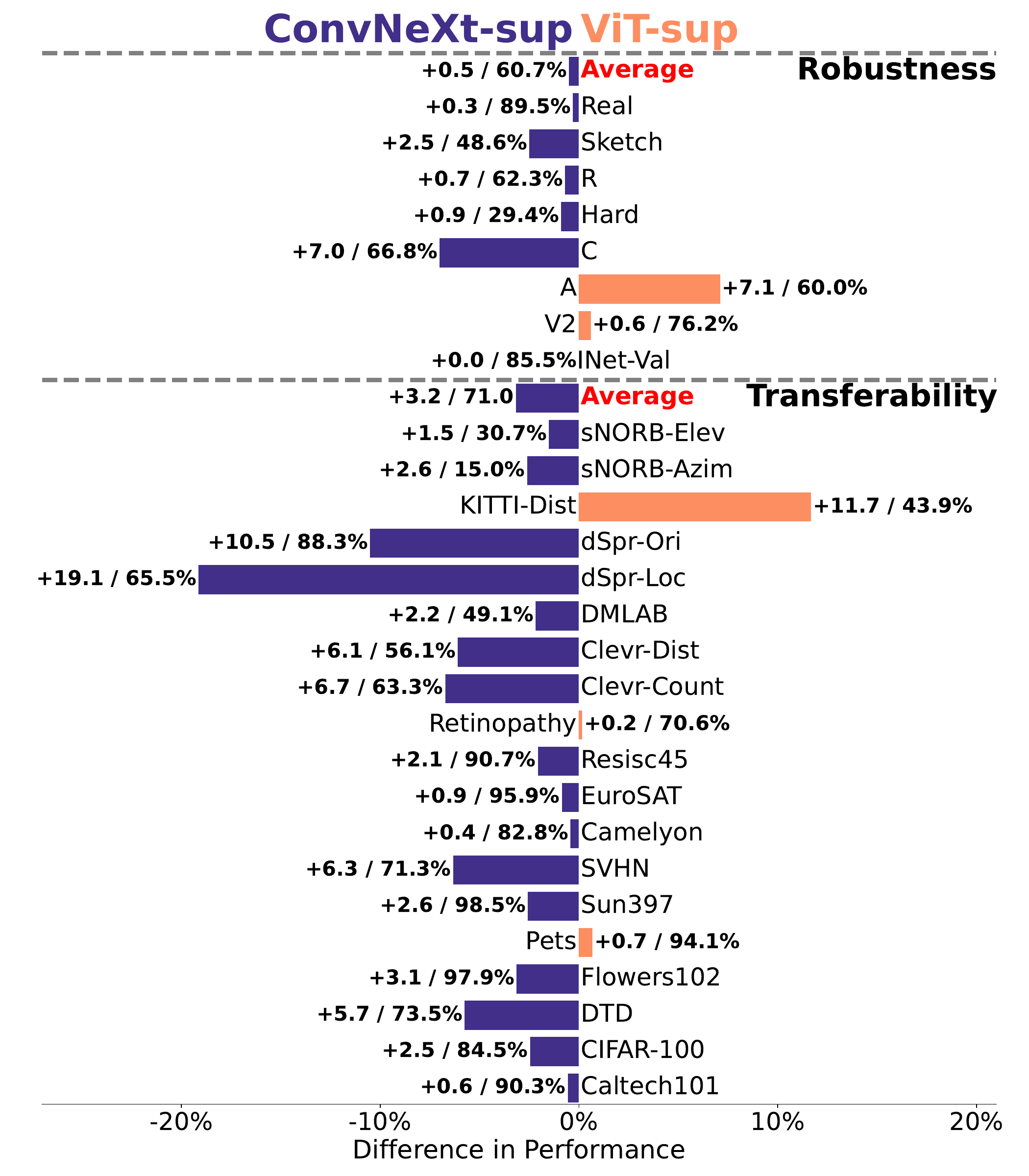}
    \end{minipage}%
    \begin{minipage}[b]{0.499\textwidth}
        \includegraphics[width=\textwidth]{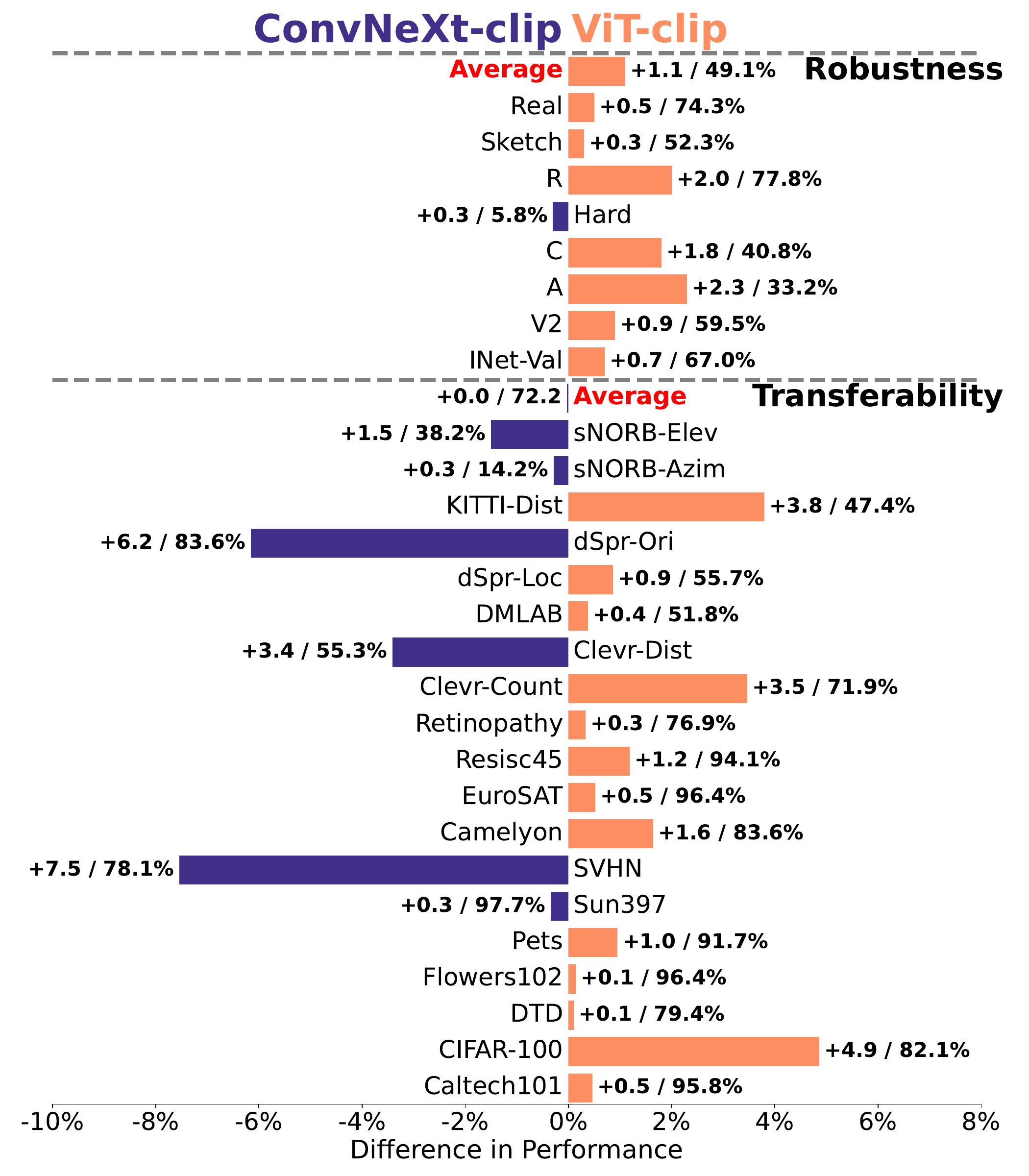}        
    \end{minipage}  
    \vspace{-0.25in}
    \caption{\textbf{Robustness (top) and transferability (bottom) results.} CLIP models excel in transferability, while supervised models are better on robustness benchmarks. In transferability, supervised ConvNeXt outperforms supervised ViT and is close to CLIP models.}
    \label{fig:vtab_robust_diff}
     \vspace{-0.1in}
\end{figure*}

\textbf{Supervised ConvNeXt is better calibrated than supervised ViT.} Our experiments reveal that supervised ConvNeXt outperforms Transformer in calibration, contrary to~\cite{minderer2021revisiting}'s findings on ViTs and ConvNets. This discrepancy is because~\cite{minderer2021revisiting} focused on older ConvNet architectures, such as ResNet, while we use a more modern one. For CLIP models, we find that ViT is only slightly better than ConvNeXt.

\subsection{Robustness}
\label{section:robustness}

A model may excel on data from its training distribution but struggle to generalize to a distribution shift~\citep{recht2019imagenet}. These shifts can arise from natural perturbations such as atmospheric conditions (e.g., fog, rain), camera noise, or variations in object location and orientation. Model robustness quantifies a model's capability to adapt to changes in data distributions. A robust model should maintain high accuracy with these perturbations. This is particularly important for applications where reliability is a primary concern. 

We evaluate the robustness on several ImageNet variants that feature many types of natural variations and corruptions: V2~\citep{recht2019imagenet}, A~\citep{hendrycks2021natural}, C~\citep{hendrycks2019benchmarking}, R~\citep{hendrycks2021many}, Sketch~\citep{wang2019learning}, Real~\citep{beyer2020we}, and Hard~\citep{taesiriimagenet}. We also provide ImageNet-1K validation accuracy for reference (INet-val). The results are shown in Fig.~\ref{fig:vtab_robust_diff} (top).

\textbf{Supervised models are better than CLIP on most of the robustness benchmarks.} In Fig.~\ref{fig:vtab_robust_diff}, we can see that supervised models perform better than CLIP on most datasets except ImageNet-R and ImageNet-Sketch. CLIP models' success on ImageNet-R and ImageNet-Sketch suggests they handle abstract or creative visuals better than supervised models. The advantage of supervised models is likely related to the fact that all robustness datasets share the same set of classes as the original ImageNet-1K, on which they were finetuned. This underscores the need for the development of new robustness benchmarks that are not directly related to ImageNet. Additionally, CLIP models may achieve higher performance when pretrained on larger datasets~\citep{fang2023data, gadre2023datacomp}. ViT and ConvNeXt, on average, have similar performance across both supervised and CLIP.

\vspace*{-0.05in}
\subsection{Transferability}
\label{section:transferability}

\begin{figure*}[!htbp]
  \begin{minipage}[b]{0.99\textwidth}
    \centering
    \includegraphics[width=\textwidth]{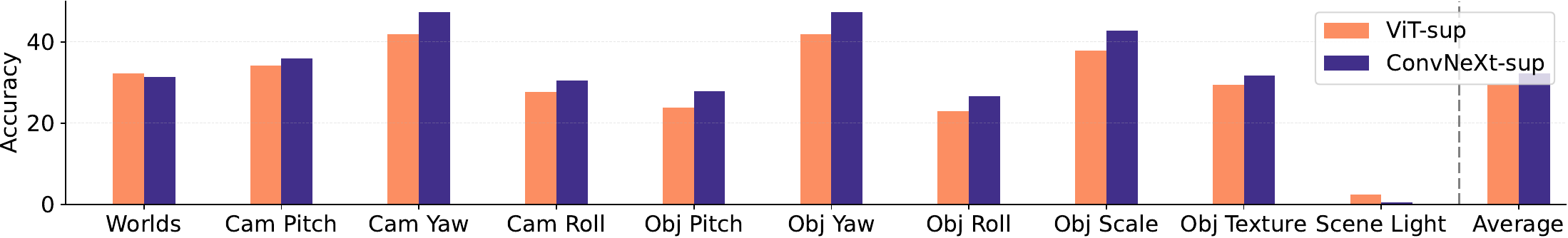}
  \end{minipage}
  \begin{minipage}[b]{0.99\textwidth}
    \centering
    \includegraphics[width=\textwidth]{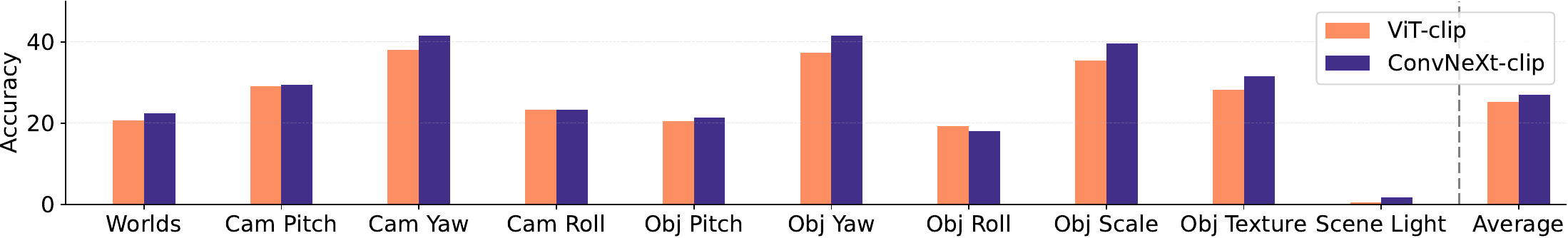}
  \end{minipage}
  \caption{\textbf{Results on synthetic data from PUG-ImageNet.} ConvNeXt is superior on almost every factor for both supervised and CLIP.}
  \label{fig:pug_imagenet}
\end{figure*}
The transfer learning performance of a model indicates its ability to adapt to new tasks and datasets beyond its original training domain~\citep{kolesnikov2020big}. Good transferability allows for rapid finetuning with minimal additional effort, making it easier to scale the model to a wide range of real-world applications. The ability of a model to adapt to these shifts without significant degradation in performance serves as a valuable metric for its utility and generalization capabilities. For instance, consider a model that has been originally trained on ImageNet, which primarily consists of natural images. A test of its transferability would be to evaluate how well this model performs when applied to a vastly different domain, such as medical imaging.

To assess the transferability, we adopted a VTAB benchmark~\citep{zhai2019large}.  It comprises 19 diverse datasets grouped into three subcategories: natural, specialized, and structured. We conduct a linear probing evaluation on frozen features, following the protocol from~\cite{ilharco_gabriel_2021_5143773}. The results are shown in Fig.~\ref{fig:vtab_robust_diff} (bottom) and Table~\ref{table:vtab}. %Transferability results on VTAB, grouped by subcategories are provided in Table~\ref{table:vtab}.

\begin{figure}[!htbp]
% \vspace{-0.1in}
\scriptsize
\def\arraystretch{1.2}
\centering
% \vspace{-0.1in}
\setlength{\tabcolsep}{5pt} % Set the horizontal padding between columns, modify 5pt to your needs
\resizebox{0.45\textwidth}{!}{ % Resize the table, set width and height accordingly
\begin{tabular}{lcccc}
\bf Model    & \bf   Natural & \bf   Specialized & \bf  Structured &  \bf  Overall \\ \shline
\vitsup  &    84.2   &     84.2      &    45.4     &  67.8   \\
\cnnsup  &    87.1   &     85.0      &    50.0     &  71.0   \\
\vitclip &    87.6   &     87.8      &    50.9     &  72.2   \\
\cnnclip &    87.8   &     86.9      &    51.2     &  72.2   \\
\end{tabular}
} % end of resizebox
\captionof{table}{\textbf{Transferability results on VTAB in subgroups.} CLIP models are better on each of the dataset subgroups. For supervised models, ConvNeXt outperforms ViT by a large margin.}
\label{table:vtab}
% \vspace{-0.1in}
\end{figure}

\textbf{Supervised ConvNeXt has great transferability, almost matching the performance of CLIP models.} We find that ConvNeXt strongly outperforms ViT for supervised. Interestingly the performance of supervised ConvNeXt is not very far from CLIP models, both of which have the same average accuracy. For CLIP, ViT and ConvNeXt demonstrate similar average accuracy, with many datasets showing a performance gap of less than 1\%. CLIP models generally show better transferability on all three subgroups of VTAB (Table \ref{table:vtab}), which is different from the robustness experiments. The superiority of CLIP can be attributed to the larger and more diverse volume of pretraining data~\citep{ramanujan2023connection}.

\subsection{Synthetic Data}

While two previous sections focused on robustness and transferability, they did not cover the new and promising area of training models with synthetic data~\citep{tian2023stablerep}. Unlike human-annotated data, synthetic datasets allow precise control over the content and quality of data.

PUG-ImageNet~\citep{bordes2023pug} is a synthetic dataset of photorealistic images of ImageNet classes that provides labels for a set of factors. The images are generated using a software that allows systematically varying factors like pose, size, texture, and others for each object. In our experiments, we provide top-1 accuracy results for ten different factors in PUG-ImageNet and their average in Fig.~\ref{fig:pug_imagenet}.

\textbf{ConvNeXt is better than ViT on synthetic data.} Intriguingly, ConvNeXt outperforms ViT on PUG-ImageNet for nearly all factors. This suggests: ConvNeXt is better than ViT on synthetic data. CLIP models have lower accuracy compared to supervised, which is likely related to their inferior performance on the original ImageNet.

\begin{figure*}[!htbp]
    \includegraphics[width=0.99\linewidth]{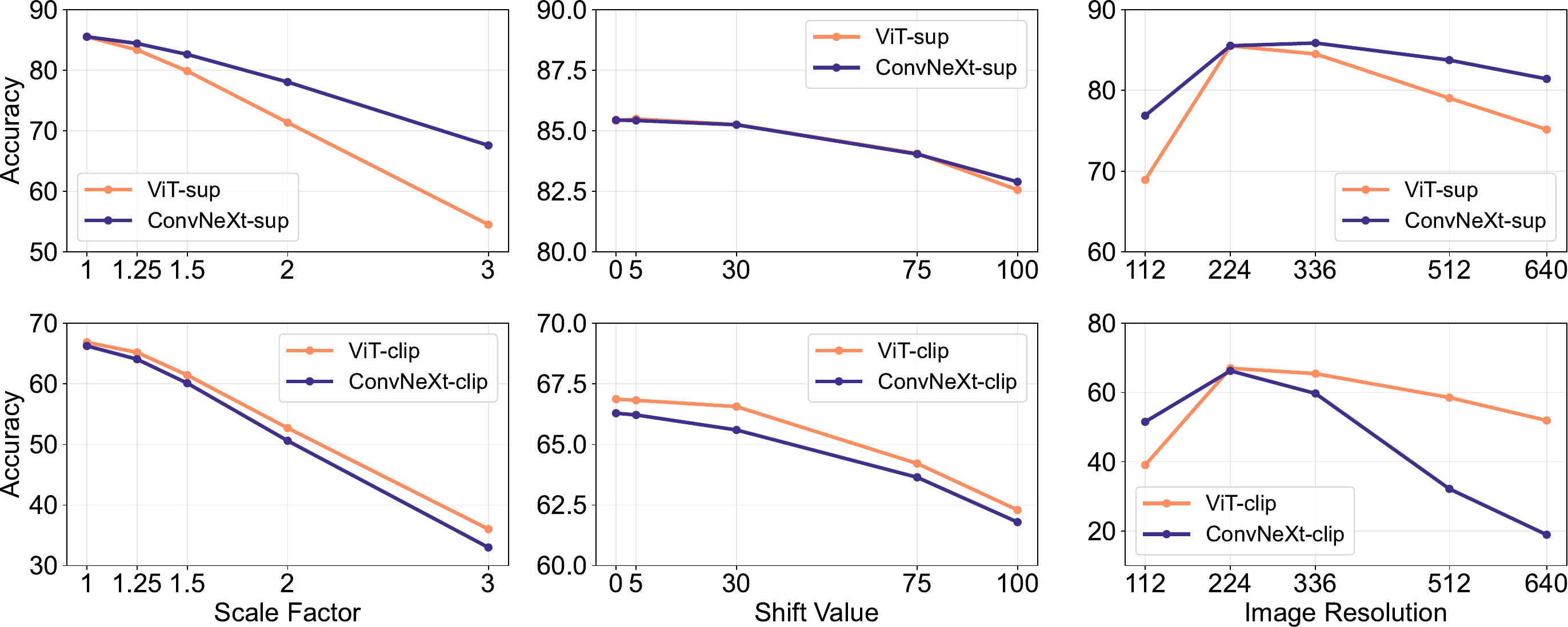}       
    \vspace{-0.1in}
    \caption{\textbf{Scale, shift, and resolution invariance experiments.} ConvNeXt is better than ViT under supervised training on all transformation types. All models are robust to shift transformation but experience degradation when the image scale is altered. }
    \label{fig:invariance}
    \vspace{-0.125in}
\end{figure*}

\vspace*{-0.05in}
\subsection{Transformation Invariance} 

In real-world scenarios, data often undergo transformations that preserve its semantic meaning or class. We aim to ensure that the model's representations are invariant to these transformations. Achieving various types of invariance is desirable because it enables the network to generalize well across different but semantically similar inputs, thereby enhancing its robustness and predictive power. In previous literature~\citep{azulay2018deep, zhang2019making}, it has been shown that the performance of neural networks can be highly unstable even under simple input data transformations, such as shifting an input image by a few pixels.

We conduct experiments to assess three types of invariance: scale, shift, and resolution. We analyze the model's accuracy trends on the ImageNet-1K validation set as a function of varying scale / shift magnitude and image resolution. In scale invariance analysis, the image is first resized according to a given scale factor, and then a central crop is taken. In shift experiments, we adjust the crop location in the original image space and then take a crop, shifting along the longer side of the image. In resolution experiments with ViT model, we interpolate positional embeddings to match the new applied resolution. 

\textbf{Supervised ConvNeXt is the most invariant model to the data transformations.} We display our results in Fig.~\ref{fig:invariance}, observing a consistent trend of ConvNeXt outperforming ViT under supervised training. Interestingly, supervised ConvNeXt has better performance on 336 pixel resolution than on the original resolution of 224 pixels. Overall, all models are robust to shifting and less robust to scaling and resolution transforms. For practical use cases requiring high transform invariance, our results indicate that supervised ConvNeXt will be the best choice among analyzed models. 

\section{Related Work}

\textbf{Architecture analysis.} 
Several works compared ViTs and ConvNeXt from the perspective of internal representations~\cite{raghu2021vision}, synthetic data~\cite{ruiz2022finding}, transferability~\cite{zhou2021convnets}, and robustness~\cite{wang2022can, NEURIPS2021_e19347e1, pinto2022impartial, djolonga2021robustness}. Other studies included analysis of Transformer properties~\cite{naseer2021intriguing} and impact of neural network width and depth on learned representations~\cite{nguyen2020wide}. ViTs and ConvNets were also evaluated on ImageNet, showing that Transformers are more aligned with human error patterns~\cite{tuli2021convolutional}. A large variety of backbones, trained with various methods, were benchmarked in~\cite{goldblum2024battle} across a diverse set of computer vision tasks, including classification, detection, and retrieval. In contrast to studies that analyze a single property, our work extensively compares models across many, maintaining a fair comparison by evaluating models with similar ImageNet accuracies.

\textbf{Training objective analysis.}  A comprehensive analysis was conducted in \cite{walmer2023teaching}, comparing ViTs trained with supervised, self-supervised, and CLIP objectives. Analysis of the representations of models trained with supervised and self-supervised objectives was presented in~\cite{grigg2021self, gwilliam2022beyond}. Two  works~\citep{park2023self, shekhar2023objectives} focused on investigating the effect of training objective in self-supervised learning. Unlike studies emphasizing self-supervised models, our work compares supervised and CLIP models.

\textbf{Limitations of ImageNet.}
Recent research~\citep{beyer2020we, recht2019imagenet, pmlr-v119-tsipras20a, yun2021re} highlighted issues with the reliability and quality of ImageNet labels. Two studies~\citep{kornblith2019better, pmlr-v139-miller21b} showed a strong relationship between performance on ImageNet and other datasets, although this can depend on the model's architecture and training methods. 
Other works~\citep{richards2023does, fang2023does} showed that high ImageNet accuracy does not guarantee good performance on diverse datasets. Current robustification training techniques were found to overfit~\citep{yamada2022does} to ImageNet evaluations. In addition, ImageNet suffers from dichotomous data difficulty~\citep{meding2021imagenet}, obscuring differences between models. Our analysis does not directly address data-related problems of ImageNet but instead studies alternative properties.

\vspace*{-0.1in}
\section{Conclusion}

Our study examined ConvNets and Transformers with supervised and CLIP training from multiple perspectives beyond the standard ImageNet accuracy. We found that models with similar ImageNet accuracies have vastly different properties. This suggests that model selection should depend on the target use cases, as standard metrics may overlook key nuances. In addition, it is crucial to develop new benchmarks with data distributions that closely mirror real-world scenarios. This will help both in training models for better real-world performance and in more accurately evaluating their effectiveness in such environments.

\textbf{ConvNet vs Transformer.} (1) Supervised ConvNeXt is superior to supervised ViT: it is more invariant to data transformations, and demonstrates better transferability, robustness and calibration. (2) ConvNeXt outperforms ViT on synthetic data. (3) ViT has a higher shape bias. 
 
 \textbf{Supervised vs CLIP.} (1) Supervised ConvNeXt competes well with CLIP in transferability, showing potential of supervised models. (2) Supervised models are better at robustness benchmarks, likely because these are ImageNet variants. (3) CLIP models have a higher shape bias and make less classification errors relative to their ImageNet accuracy.
 
As a result of our analysis, we suggest using supervised ConvNeXt when the target task distribution is not very different from ImageNet as this model provides competitive performance among many benchmarks. In case of a serious domain shift, we recommend using CLIP models.

\textbf{Acknowledgments.} We would like to thank Sheng Zhang, Kudaibergen Abutalip, Guangyi Chen, Nurdaulet Mukhituly, Boyang Sun, Zhiqiu Xu, and Nurbek Tastan for their valuable feedback and suggestions. Use of all datasets and all experiments conducted by MBZUAI.

\section*{Impact Statement}

This paper presents work whose goal is to advance the fields of Machine Learning, Deep Learning and Computer Vision. There are many potential societal consequences of our work, none which we feel must be specifically highlighted here.

\nocite{langley00}

\bibliography{example_paper}
\bibliographystyle{icml2024}

\newpage
\appendix
\label{appendix}
\onecolumn

\section*{\Large{Appendix}}
\vspace{0.5em}
\section{Results for Models of Different Sizes}
\label{appendix:models}
Besides having the base sized models in the main part of our analysis, we additionally include the following models to see the effect of the model size on the performance:\begin{itemize}
\addtolength{\itemsep}{-0.05in}
\item Supervised ConvNeXt: Tiny, Small, Large, Huge (XLarge).
\item Supervised ViT (DeiT 3): Small, Large, Huge (XLarge). Note that no Tiny version is provided for DeiT 3 by the original authors~\citep{touvron2022deit}.
\item CLIP ConvNeXt: Large, Huge (XLarge).
\item CLIP ViT: Large, Huge.
\end{itemize}
All new additional CLIP models are pretrained on the LAION-2B dataset, while the CLIP models from the main part of the manuscript (Table~\ref{tab:model_details}) are pretrained on LAION-400M. All CLIP models are taken from OpenCLIP, and supervised models are taken from the respective original works~\citep{liu2022convnet, touvron2022deit}. 

\begin{table}[h]
\centering
\small
\addtolength{\tabcolsep}{0pt}
\def\arraystretch{1.2}
\label{table:model-configurations}
\begin{tabular}{lcccccc}
\textbf{Architecture} & \textbf{Pretraining} & \textbf{Finetuning} & \textbf{Paradigm} & \textbf{FLOPs} & \textbf{\#Params} & \textbf{INet-1K val\%} \\
\shline
ConvNeXt-Tiny & ImageNet-21K & ImageNet-1K & Supervised & 4.5G & 29M & 82.9 \\
ConvNeXt-Small & ImageNet-21K & ImageNet-1K & Supervised & 8.7G & 50M	& 84.6 \\
ConvNeXt-Large & ImageNet-21K & ImageNet-1K & Supervised & 34.4G & 198M & 86.6 \\
ConvNeXt-Huge & ImageNet-21K & ImageNet-1K & Supervised & 60.9G & 350M & 87.0 \\
ViT-S/16 & ImageNet-21K & ImageNet-1K & Supervised & 4.6G & 22M & 83.1 \\
ViT-L/16 & ImageNet-21K & ImageNet-1K & Supervised & 61.6G & 304M	 & 87.0 \\
ViT-H/14 & ImageNet-21K & ImageNet-1K & Supervised & 167.4G & 632M & 87.2 \\
ConvNeXt-Large & LAION-2B & --- & CLIP & 34.4G & 198M & 75.2 \\
ConvNeXt-Huge & LAION-2B & --- & CLIP & 60.9G & 350M & 78.6 \\
ViT-L/14 & LAION-2B & --- & CLIP & 61.6G & 304M & 74.8 \\
ViT-H/14 & LAION-2B & --- & CLIP & 167.4G & 632M & 77.3 \\
\end{tabular}
\caption{\textbf{Additional models configurations.} We perform partial analysis for the additional models of different sizes to see the effect of the model size on the performance.}
\end{table}

Additional models are evaluated on PUG-ImageNet (Table~\ref{table:pug-imagenet-appendix}), ImageNet-X (Table~\ref{table:imagenet-x-appendix}), calibration (Table~\ref{table:calibration-1k-appendix} and Table~\ref{table:calibration-r-appendix}), transformation invariance (Table~\ref{table:invariance-appendix}), and shape / texture bias (Table~\ref{table:shape-texture-bias-appendix}). Based on the results from new models we make the following observations:
\begin{itemize}
\addtolength{\itemsep}{-0.05in}
\item On PUG-ImageNet (Table~\ref{table:pug-imagenet-appendix}) ConvNeXt is better than ViT in 6 out of 7 comparison, suggesting its clear advantage over ViT on synthetic data.
\item ImageNet-X performance is largely determined by the training method (Table 5). CLIP models are clearly better than supervised.
\item On calibration (Table~\ref{table:calibration-1k-appendix} and Table~\ref{table:calibration-r-appendix}) ConvNeXt has lower ECE value compared to its ViT counterpart in most cases. This solidifies our initial conclusion in the main part that ConvNeXt is better calibrated than ViT.
\item Shape bias greatly improves with scale (Table \ref{table:shape-texture-bias-appendix}). Interestingly, the ConvNeXt is better than ViT on large-scale CLIP models (Large and Huge). For Huge CLIP models ConvNeXt has an advantage of almost 10\% over ViT. This suggests that training method and model size has a noticeable influence on the shape bias of the model. Moreover, this result also highlights that ConvNeXt has the potential to exceed a ViT on this benchmark.
\item Even small supervised models are robust to shift transformation (Table \ref{table:invariance-appendix} middle part). For example, the smallest supervised model ConvNeXt-Tiny has a small degradation of $<3\%$ which is comparable to the huge CLIP models.
\item Huge supervised models are quite reliable to the resolution change (Table \ref{table:invariance-appendix} right part).
\item In general, large models provide a decent improvement over the base models. However, huge models provide only marginal improvement over the large models.
\end{itemize}

Overall, we observe that the conclusions from the Base-sized models largely transfer to other model sizes. This includes the superiority of ConvNeXt over ViT on synthetic data, the dependence of ImageNet-X performance on the training method, better calibration of ConvNeXt, greater shape bias of supervised ViT, and better robustness of supervised ConvNeXt compared to supervised ViT.

\begin{table}[h]
\centering
\begin{minipage}[t]{0.48\textwidth}
\centering
\small
\def\arraystretch{1.2}
\begin{tabular}{lccccc}
\textbf{Model} & \textbf{Tiny} & \textbf{Small} & \textbf{Base} & \textbf{Large} & \textbf{Huge} \\
\shline
ConvNeXt-sup & 27.0 & 29.4 & 32.2 & 34.6 & 35.6 \\
ViT-sup & — & 24.7 & 29.4 & 34.3 & 36.9 \\
ConvNeXt-clip & — & — & 27.0 & 35.4 & 43.2 \\
ViT-clip & — & — & 25.2 & 32.9 & 38.8 \\
\end{tabular}
\caption{\textbf{PUG-ImageNet results.} ConvNeXt is better in 6 out of 7 comparisons.}
\label{table:pug-imagenet-appendix}
\end{minipage}\hfill
\begin{minipage}[t]{0.48\textwidth}
\centering
\small
\def\arraystretch{1.2}
\begin{tabular}{lccccc}
\textbf{Model} & \textbf{Tiny} & \textbf{Small} & \textbf{Base} & \textbf{Large} & \textbf{Huge} \\
\shline
ConvNeXt-sup & 1.48 & 1.50 & 1.43 & 1.47 & 1.50 \\
ViT-sup & — & 1.46 & 1.45 & 1.49 & 1.50 \\
ConvNeXt-clip & — & — & 1.23 & 1.26 & 1.32 \\
ViT-clip & — & — & 1.19 & 1.33 & 1.31 \\
\end{tabular}
\caption{\textbf{ImageNet-X results.} CLIP models have an advantage.}
\label{table:imagenet-x-appendix}
\end{minipage}
\end{table}

\begin{table}[h]
\begin{minipage}{0.48\textwidth}
    \centering
    \small
    \def\arraystretch{1.2}
    \begin{tabular}{lccccc}
        \textbf{Model} & \textbf{Tiny} & \textbf{Small} & \textbf{Base} & \textbf{Large} & \textbf{Huge} \\
        \shline
        ConvNeXt-sup & 0.021 & 0.022 & 0.025 & 0.029 & 0.031 \\
        ViT-sup & — & 0.034 & 0.043 & 0.047 & 0.041 \\
        ConvNeXt-clip & — & — & 0.085 & 0.017 & 0.019 \\
        ViT-clip & — & — & 0.073 & 0.030 & 0.031 \\
    \end{tabular}
    \captionof{table}{\textbf{Calibration on ImageNet-1K}. ConvNeXt has lower ECE in most cases.}
    \label{table:calibration-1k-appendix}
\end{minipage} \hfill
\begin{minipage}{0.48\textwidth}
    \centering
    \small
    \def\arraystretch{1.2}
    \begin{tabular}{lccccc}
        \textbf{Model} & \textbf{Tiny} & \textbf{Small} & \textbf{Base} & \textbf{Large} & \textbf{Huge} \\
        \shline
        ConvNeXt-sup & 0.073 & 0.035 & 0.028 & 0.022 & 0.023 \\
        ViT-sup & — & 0.056 & 0.041 & 0.037 & 0.037 \\
        ConvNeXt-clip & — & — & 0.141 & 0.080 & 0.090 \\
        ViT-clip & — & — & 0.133 & 0.094 & 0.098 \\
    \end{tabular}
    \captionof{table}{\textbf{Calibration on ImageNet-R.} ConvNeXt is better calibrated than ViT.}
    \label{table:calibration-r-appendix}
\end{minipage}
\end{table}

\begin{table}[!ht]
    \scriptsize
\def\arraystretch{1.2}
\centering 
    \begin{tabular}{lccccc|ccccc|ccccc}
    \bf Model & \multicolumn{5}{c|}{\bf Scale Invariance} & \multicolumn{5}{c|}{\bf Shift Invariance} & \multicolumn{5}{c}{\bf Resolution Invariance} \\ 
         & \bf 1x & \bf 1.25x & \bf 1.5x & \bf 2x & \bf 3x & \bf 0px & \bf 5px & \bf 30px & \bf 75px & \bf 100px & \bf 112px & \bf 224px & \bf 336px & \bf 512px & \bf 640px \\ \shline 
        ConvNeXt-sup Tiny & 82.9 & 81.8 & 79.5 & 74.6 & 63.2 & 82.9 & 82.9 & 82.6 & 81.4 & 80.1 & 67.2 & 82.9 & 77.2 & 42.8 & 21.1 \\ 
        ConvNeXt-sup Small & 84.6 & 83.3 & 81.4 & 76.8 & 66.4 & 84.6 & 84.5 & 84.5 & 83.3 & 82.1 & 74.8 & 84.6 & 84.7 & 82.5 & 80.6 \\ 
        ConvNeXt-sup Base & 85.5 & 84.5 & 82.7 & 78.3 & 68.5 & 85.5 & 85.7 & 85.6 & 84.5 & 83.2 & 78.1 & 85.5 & 85.8 & 83.7 & 81.5 \\ 
        ConvNeXt-sup Large & 86.6 & 85.4 & 84.9 & 79.9 & 70.7 & 86.6 & 86.6 & 86.5 & 85.4 & 84.5 & 80.2 & 86.6 & 86.2 & 84.6 & 82.9 \\ 
        ConvNeXt-sup Huge & 87.0 & 85.8 & 84.3 & 80.4 & 71.5 & 87.0 & 86.8 & 86.7 & 85.7 & 84.7 & 81.0 & 87.0 & 86.7 & 84.8 & 82.9 \\ \hline
        ViT-sup Small & 82.7 & 80.5 & 76.4 & 65.7 & 48.2 & 82.7 & 82.8 & 82.5 & 81.1 & 79.2 & 65.5 & 82.7 & 81.7 & 75.4 & 70.9 \\
        ViT-sup Base & 85.5 & 83.3 & 79.7 & 70.6 & 54.8 & 85.5 & 85.5 & 85.2 & 84.1 & 82.6 & 69.5 & 85.5 & 84.2 & 78.1 & 74.5 \\ 
        ViT-sup Large & 86.8 & 85.0 & 81.9 & 74.2 & 60.1 & 86.8 & 86.8 & 86.6 & 85.5 & 84.3 & 76.7 & 86.8 & 85.9 & 81.7 & 79.1 \\ 
        ViT-sup Huge & 86.9 & 85.2 & 81.8 & 73.9 & 59.6 & 86.9 & 86.8 & 86.7 & 85.7 & 84.5 & 73.1 & 86.9 & 85.9 & 81.7 & 79.4 \\ \hline
        ConvNeXt-clip Base & 66.3 & 64.0 & 60.0 & 50.4 & 33.3 & 66.3 & 66.2 & 65.7 & 63.6 & 61.8 & 51.5 & 66.3 & 59.7 & 32.2 & 18.9 \\ 
        ConvNeXt-clip Large & 75.2 & 73.5 & 71.3 & 64.9 & 50.6 & 75.2 & 75.1 & 74.8 & 73.5 & 72.1 & 62.1 & 75.2 & 76.1 & 70.3 & 64.3 \\ 
        ConvNeXt-clip Huge & 78.6 & 77.3 & 75.2 & 69.6 & 56.3 & 78.6 & 78.6 & 78.2 & 77.1 & 75.8 & 68.0 & 78.6 & 78.8 & 71.0 & 61.9 \\ \hline
        ViT-clip Base & 67.0 & 65.0 & 61.3 & 52.5 & 36.1 & 67.0 & 66.9 & 66.6 & 64.2 & 62.2 & 39.1 & 67.0 & 65.4 & 58.5 & 51.9 \\ 
        ViT-clip Large & 74.8 & 72.1 & 68.1 & 58.7 & 41.4 & 74.8 & 74.7 & 74.3 & 72.0 & 69.6 & 48.0 & 74.8 & 74.4 & 69.5 & 64.9 \\ 
        ViT-clip Huge & 77.3 & 74.9 & 71.4 & 62.7 & 45.5 & 77.3 & 77.4 & 76.9 & 74.7 & 72.7 & 51.1 & 77.3 & 76.9 & 72.7 & 68.4 \\ 
    \end{tabular}
    \caption{\textbf{Scale, shift, and resolution invariance experiments.}}
    \label{table:invariance-appendix}

\end{table}

\begin{table}[H]
\centering
\small
\def\arraystretch{1.2}
\centering
\begin{tabular}{lccccc}
\textbf{Model} & \textbf{Tiny} & \textbf{Small} & \textbf{Base} & \textbf{Large} & \textbf{Huge} \\
\shline
ConvNeXt-sup & 28.1 & 31.6 & 33.3 & 39.8 & 41.3 \\
ViT-sup & — & 35.0 & 39.8 & 51.3 & 57.1 \\
ConvNeXt-clip & — & — & 45.9 & 56.9 & 67.0 \\
ViT-clip & — & — & 46.4 & 53.9 & 58.5 \\
\end{tabular}
\caption{\textbf{Shape-texture bias results.} Shape bias noticeably improves with model size.}
\label{table:shape-texture-bias-appendix}
\end{table}

\end{document}